\documentclass[12pt]{article}

\usepackage{mrfcommon}

\title{An Approximation Method for Fitted Random Forests}
\author{}
\author[$1$]{Sai K. Popuri\thanks{For Correspondence: \url{sai.popuri@gmail.com}}}
\date{}

\begin{document}
\maketitle

\begin{abstract}
Random Forests (RF) is a popular machine learning method for classification and regression problems. It involves a bagging application to decision tree models. One of the primary advantages of the Random Forests model is the reduction in the variance of the forecast. In large scale applications of the model with millions of data points and hundreds of features, the size of the fitted objects can get very large and reach the limits on the available space in production setups, depending on the number and depth of the trees. This could be especially challenging when trained models need to be downloaded on-demand to small devices with limited memory. There is a need to approximate the trained RF models to significantly reduce the model size without losing too much of prediction accuracy. In this project we study methods that approximate each fitted tree in the Random Forests model using the multinomial allocation of the data points to the leafs. Specifically, we begin by studying whether fitting a multinomial logistic regression (and subsequently, a generalized additive model (GAM) extension) to the output of each tree helps reduce the size while preserving the prediction quality. 

\end{abstract}

\section{Problem Statement}
\label{sec:intro} 

Suppose we have the data $\mathscr{D}=$ $(y_i, \mathbf{x}_i)$, $i=1,\ldots,N$, where $y_i$ is continuous and $\mathbf{x}_i$ is a $p$-dimensional vector of features, which can be continuous, discrete, or a mix of both. A Random Forests model can be fitted to the data 
$\mathscr{D}$ by first building $M$ sets of samples of size $n \leq N$. These samples are selected without replacement from the data. To each sample, $\mathscr{D}_m$ a decision tree is fitted to get a tree object $T(\mathscr{D}_m)$. The forecast for a new set of features $\mathbf{x}^*$ is given by
\begin{equation}
\label{eq:rf-forecast}
\hat{y}(\mathbf{x}^*) = \frac{1}{M}\sum_{m=1}^{M}\hat{f}(\mathbf{x}^*, T(\mathscr{D}_m)),
\end{equation}
where $\hat{f}$ is a summary (e.g.: average, median etc.) of the responses $y_j$s allocated to the leaf traversed to by the tree $T$ when $\mathbf{x}^*$ is given as the input. Some of the important parameters to Random Forests are $n$, $k \leq p$, $d$, that determines how deep the trees are to be grown, and $M$, the number of trees. When given the parameter $k$, the model randomly selects $k$ features out of $p$ at each split decision. To keep notation simple, let us denote the Random Forest model by $R(n, k, d, M)$.

In most commercial production systems, the models are first trained and the resulting fitted or trained objects are stored for subsequent retrieval for scoring. In large scale applications, where millions of models need to be fitted, the size of the fitted models is of importance. In Random Forests, the fitted trees $T(\mathscr{D}_m)$s, along with some book-keeping information is stored as an object $\hat{R}(n, k, d, M)$, where $\hat{R}$ is the fitted model. Sometimes the size of the fitted trees $T(\mathscr{D}_m)$s and therefore the size of the fitted Random Forests model $\hat{R}(n, k, d, M)$ increase beyond some acceptable levels. One way to reduce the size of the fitted objects is by adjusting the parameter values $(k, n, d, M)$. While this does help reduce the size, the model also suffers loss of prediction accuracy. In this research, we are interested in finding ways to approximate $\hat{R}(n, k, d, M)$ with say, $\tilde{R}(n, k, d, M)$ so that the size reduction vs prediction accuracy trade-off is better than changing the parameters. Specifically, we begin by approximating the regression fit in the trees $T(\mathscr{D}_m)$s with a multinomial logistic regression. 

\subsection{Multinomial logistic regression}
Let us briefly review the logistic regression\citep{ESL2009}. Suppose we have the binary response data $y_i$ that take one of the two categories (typically, coded as $0$ or $1$) as the response. To each $y_i$, we have a set of features or covariates $\mathbf{x}_i$, which is vector of dimension $p$. The logistic regression model assumes that each $Y_i$ is Bernoulli distributed with parameter $\text{Pr}(Y_i = 1) = \theta_i$ and parameterizes $\theta_i$ using the logit function 
\begin{equation}
\label{eq:logit-fn}
\log\frac{\theta_i}{1-\theta_i} = \alpha + \mathbf{x}^{T}_i\mathbf{\beta},
\end{equation}
where $\alpha$ and $\mathbf{\beta}$ are the parameters that need to estimated. The estimation is typically done by maximizing the likelihood function using a numerical method like Newton-Raphson or Fisher scoring. The multinomial logistic regression is an extension of the logistic regression. Let us first look at the multinomial distribution. The multinomial distribution is a natural extension of the Binomial distribution when each independent trial has more than two possible mutually exclusive outcomes. Consider a series of $n$ independent trials, each resulting in one of the $(m+1)$ mutually exclusive events $E_1,\ldots,E_{m+1}$. In each trial, suppose the probability of occurrence of event $E_i$ is equal to $\theta_i$, with 
$\sum_{i=1}^{m+1}\theta_i = 1$. Let $\mathbf{X} = (X_1,\ldots,X_{m+1})^{T}$ denote the random vector of the number of occurrences of events $E_1,\ldots,E_{m+1}$ out of $n$ trials, with $\sum_{i=1}^{m+1}X_i = n$. Let $\mathbf{x} = (x_1,\ldots,x_{m+1})^{T}$ represent a realization of $\mathbf{X}$, $\sum_{i=1}^{m+1}x_i = n$. Then, the random vector $\mathbf{X}$ is said to have a Multinomial distribution with parameters $(\mathbf{\theta}; n)$ with $\mathbf{\theta} = (\theta_1,\ldots,\theta_{m+1})^{T}$. The joint probability density function of $\mathbf{X}$ is given by 
\begin{equation}
\label{eq:multinomialpdf}
\text{Pr}(\mathbf{X}=\mathbf{x}) = \frac{n!}{x_1!x_2!\ldots x_{m+1}!}\prod_{i=1}^{m+1}\theta_i^{x_i}.
\end{equation}
Since $n$ is known, $X_{m+1}$ and $\theta_{m+1}$ do not provide any additional information since $X_{m+1}=n-\sum_{i=1}^{m}X_i$ and $\theta_{m+1} = 1-\sum_{i=1}^{m}\theta_i$. Therefore, we can reduce the dimensionality of $\mathbf{X}$ and $\mathbf{\theta}$ by deleting their respective last elements and define $\mathbf{X}=(X_1,\ldots,X_m)^{T}$ and $\mathbf{\theta} = (\theta_1,\ldots,\theta_m)^{T}$. Thus, without loss of generality, we say that $\mathbf{X}$ has a Multinomial distribution with parameters $(\mathbf{\theta}; n)$ with joint probability distribution as in \eqref{eq:multinomialpdf} with $x_{m+1}=n-\sum_{i=1}^{m}x_i$ and $\theta_{m+1} = 1-\sum_{i=1}^{m}\theta_i$. Suppose the response $Y_i$ is from a multinomial distribution with $n=1$. If we model the parameters $\theta_i$ in terms of the $\log$ of odds relative to an arbitrarily chosen base category as in \eqref{eq:multinomial-logit-fn}, we get the multinomial logistic regression\citep{multinomial-logit}. 
\begin{equation}
\label{eq:multinomial-logit-fn}
\log\frac{\theta_{ij}}{\theta_{iJ}} = \alpha_j + \mathbf{x}^{T}_i\mathbf{\beta}_j.
\end{equation}
Notice that each category $j$ has its own set of regression parameters $\alpha_j$ and $\mathbf{\beta}_j$. Estimation of the parameters is performed in a similar fashion as in the logistic regression. 

\subsection{Multinomial interpretation in a decision tree}
Consider a single decision tree $T(\mathscr{D}_m)$ fitted to a sample of size $n$. Suppose all the $p$ features were used to fit the tree. Further suppose the tree has $K$ number of leafs. This means that each data point $(Y_i, \mathbf{x}_i)$ has been allocated to one and only one of the $K$ leafs. This setup immediately lends itself to a multinomial logistic regression model with the $K$ categories treated as `nominal', i.e., there is no ordering among the categories. We fit a multinomial logistic regression to this representation to get the fitted object $U(\mathscr{D}_m; T)$. We index the model fit $U$ with the specific tree $T(\mathscr{D}_m)$ to highlight that the multinomial data is derived from the fitted decision tree $T(\mathscr{D}_m)$. 

Substituting $U$ for $T$ in \eqref{eq:rf-forecast}, we get 
\begin{equation}
\label{eq:mrf-forecast}
\tilde{y}(\mathbf{x}^*) = \frac{1}{M}\sum_{m=1}^{M}\tilde{f}(\mathbf{x}^*, U(\mathscr{D}_m; T)),
\end{equation}
where $\tilde{f}$ now predicts using the fitted multinomial logistic model instead of the decision tree $T$. Notice that even though we specify $T$ as a parameter to $U$, we do not need to store the object $T$. Suppose the resulting model object is $MR(n, k, d, M)$.

The claim is that forecasts $\tilde{y}$ are sufficiently close in accuracy to $\hat{y}$ and the size reduction in $MR(n, k, d, M)$ is greater than a size reduction in $R(n, k, d, M)$ after the parameters $(n, k, d, M)$ are changed to reduce the size of the trees.

\section{Numerical Examples}
\label{sec:application}
\begin{enumerate}
	\item Example 1: A popular Random Forest example from literature.
	\item Example 2: Demand forecasting example.
\end{enumerate}
Compare and analyze the runtime and storage space consumed, and the prediction accuracy. Hopefully, our method will result in substantial storage space reduction while maintaining desirable prediction accuracy and runtime. 

\section{Conclusion}
\label{sec:conclusion}
 
\bibliographystyle{plainnat}
\bibliography{bibfile}

\end{document}